%% file: main.tex
\documentclass[10pt]{article}

\PassOptionsToPackage{svgnames}{xcolor}
\usepackage{style}

\usepackage[numbers,sort&compress]{natbib}
\usepackage{graphicx}
\usepackage[font=small,labelfont=bf]{caption}
\usepackage{subcaption}
\usepackage{booktabs}
\usepackage{colortbl}
\usepackage{multirow}
\usepackage{bigstrut}
\usepackage{float}
\usepackage[all]{hypcap}
\usepackage{pifont}
\usepackage{amsmath}
\usepackage{mathtools}
\usepackage{mathrsfs}
\usepackage{nicefrac}

\usepackage{algorithm}
\usepackage{algpseudocode}

\usepackage{microtype}
\usepackage{enumitem}
\usepackage{cleveref}
\usepackage{bxcoloremoji}
\usepackage{url}
\usepackage{wrapfig}
\usepackage{lipsum}
\usepackage{stackengine}
\usepackage{adjustbox}
\usepackage{rotating}
\usepackage{makecell}

\begin{document}

\Cover{
  title={TacWAM: Anchor-Guided World Action Model with Mechanics-Aware Tactile Prediction},
  authors={Lei Jin\textsuperscript{*1}, Yiding Ma\textsuperscript{*1}, Xin Zhang\textsuperscript{2}, Chen Gao\textsuperscript{1$\dagger$}, Wei Wu\textsuperscript{1,2}, Yong Li\textsuperscript{1$\dagger$}},
  affils={\textsuperscript{1}Tsinghua University \quad \textsuperscript{2}Manifold AI\\ \textsuperscript{*}Equal contribution \quad \textsuperscript{$\dagger$}Co-corresponding authors: liyong07@tsinghua.edu.cn, chgao96@gmail.com},
  abstract = {\textbf{Abstract.} World Action Models (WAMs) combine future-state prediction with robot action generation, but existing approaches largely rely on visual futures. Visual prediction captures scene structure and object motion, yet provides limited supervision for force, deformation, shear, and slip during contact-rich manipulation. This creates two design requirements: tactile futures should carry meaningful physical information, and they should not become privileged cues for action generation. We present \textbf{TacWAM}, a mechanics-aware tactile WAM that addresses this challenge in three steps. First, a Spatially Aligned Fusion (SAF) Tactile Encoder maps tactile appearance, dense force fields, and deformation flow into a shared latent prediction space, with bilateral force and torque reconstruction preserving global contact information. Second, a tactile history encoder provides temporal context so future tactile prediction reflects how force and deformation change beyond the current tactile observation. Third, Anchor-Guided Tri-Modal (AGT) Attention separates current visual and tactile anchors, future prediction tokens, and action tokens, allowing future tactile states to supervise training without being directly read by the action branch. We evaluate TacWAM on four real-world contact-rich manipulation tasks covering fragile grasping, sustained surface contact, and dynamic in-hand manipulation. TacWAM achieves an average success rate of 75.0\%, exceeding the strongest evaluated baseline by 37.5 percentage points. Staged ablations show consistent degradation when tactile history is removed and access to future prediction targets is relaxed. These results indicate that future tactile supervision can improve contact-aware action learning when combined with informative tactile representations and deployment-consistent information constraints.},
  date={August 6, 2026},
  pageurl={},
  headerlogo={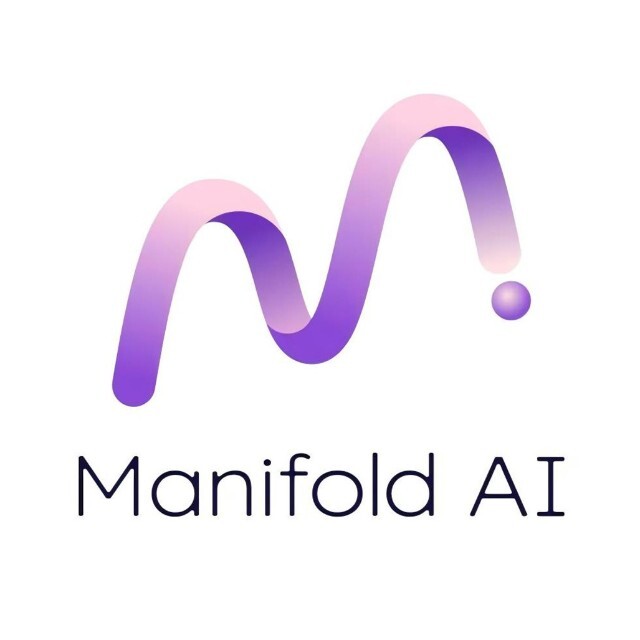},
  headerdate={August 6, 2026}
}

\input{sec/1_introduction}
\input{sec/2_related_work}
\input{sec/3_method}
\input{sec/4_experiments}
\input{sec/5_conclusion}
\bibliography{references}

\input{sec/6_appendix}

\end{document}

%% file: sec/1_introduction.tex
\section{Introduction}

World models enable robots to learn predictive representations of how the environment evolves, and recent World Action Models (WAMs) combine such predictive learning with action generation for robot control~\cite{hafner2021dreamerv2,han2026wam,liu2026oawam,cai2026ahawam}. However, most WAMs still model the future primarily through visual state transitions. Visual prediction captures scene layout, object appearance, and visible motion, but physical manipulation also depends on local mechanics that are only indirectly expressed in images. For contact-rich tasks, learning how the scene looks is therefore not sufficient; the model should also capture how tactile signals such as force and deformation change during interaction.

Tactile sensing provides direct evidence about this interaction process. Visually similar states can correspond to substantially different tactile conditions. For example, \textit{a stable-looking grasp may be close to slip, an apparently aligned insertion may already be mechanically constrained, and a brittle object may look unchanged while force becomes unsafe.} These distinctions are reflected in force buildup, deformation, shear, pressure asymmetry, and incipient slip~\cite{lepora2026tactile,ai2024robopack,zhao2024tacman,cao2026tactilefusion}. Touch is therefore useful not only for detecting contact, but also for supervising force- and deformation-related changes that visual futures alone cannot fully capture.

Prior visuo-tactile methods have used touch for observation fusion, representation learning, policy refinement, and future tactile prediction~\cite{heng2025vitacformer,zhang2025vtla,bi2025vlatouch,zang2026tacforesight,wu2026tactilewam}. These works show that touch can improve manipulation and that future tactile prediction can be useful. Building on this direction, we focus on a complementary requirement for tactile WAMs: future tactile states should provide physically informative supervision while action generation remains restricted to information available at deployment. This lets tactile futures enrich world-model training without serving as direct action inputs.

Following this principle, we introduce \textbf{TacWAM}, a tactile-augmented World Action Model that uses tactile prediction to strengthen WAM training without exposing future tactile targets to the action branch. TacWAM realizes this idea through four connected design choices. First, it maps heterogeneous gripper-side tactile observations into a shared tactile representation using a \textbf{Spatially Aligned Fusion Tactile Encoder (SAF Tactile Encoder)}, which fuses tactile appearance, dense force fields, and deformation flow, while reconstructing bilateral resultant force and torque as global wrench supervision. Second, a tactile history context summarizes recent interaction changes before the current action chunk. Third, future tactile states provide an additional predictive target for force- and deformation-aware training, serving as supervision for contact representations that visual prediction alone does not capture rather than as direct action inputs. Finally, \textbf{Anchor-Guided Tri-Modal Attention (AGT Attention)} organizes visual, tactile, and action streams around the current visual frame and current observed tactile state: future visual and tactile tokens serve as prediction targets, while action tokens can access only deployment-available anchors and action tokens. Together, these designs allow TacWAM to learn richer contact-related predictive representations during training while maintaining deployment-consistent action generation.

Our contributions can be summarized as follows:
\begin{itemize}[leftmargin=*]
    \item We introduce TacWAM, which extends visual WAM training with recent tactile history and future tactile prediction, enabling the model to learn from changes in touch beyond the current tactile observation.

    \item We develop a Spatially Aligned Fusion Tactile Encoder that maps tactile appearance, force, and deformation into a shared latent representation while preserving local contact patterns and global force/torque information.

    \item We propose Anchor-Guided Tri-Modal Attention, which prevents the action branch from accessing future visual or tactile targets during training, ensuring that action generation uses only information available at deployment.

    \item We evaluate TacWAM on four real-world contact-rich manipulation tasks. TacWAM consistently outperforms representative vision-only and visuo-tactile baselines, while staged ablations support the importance of tactile history and restricted future-target visibility in the complete framework.
\end{itemize}

%% file: sec/2_related_work.tex
\section{Related Work}

\subsection{World Action Models}

World models learn predictive representations of environment dynamics
to support planning and policy learning~\cite{hafner2021dreamerv2}.
World Action Models (WAMs) extend this idea by combining future-state
prediction with robot action generation. Recent WAMs jointly learn
actions and future visual observations or latent visual states, using
visual prediction to improve the representations learned for
control~\cite{han2026wam,liu2026oawam,cai2026ahawam}. Such prediction
captures scene structure, object motion, and other visible state
changes. However, visually similar states can involve different
contact conditions, such as stable contact, increasing force, or
incipient slip. Visual futures therefore provide limited supervision
for force, deformation, and other local contact changes. TacWAM
extends the predictive learning of WAMs by introducing future tactile
states as additional training targets.

\subsection{Visuo-Tactile and Predictive Tactile Learning}

Tactile sensing provides direct observations of contact states that
are difficult to infer from images. Prior work has learned
visuo-tactile representations for dexterous manipulation and
tactile-informed control of articulated objects
~\cite{lee2019making,heng2025vitacformer,zhao2024tacman}. Tactile information has
also been incorporated into vision-language-action policies through
multimodal features, tactile feedback pathways, and contact-aware
policy adaptation
~\cite{zhang2025vtla,huang2025tactilevla,bi2025vlatouch}. Reactive
tactile policies further use high-frequency feedback to correct
actions during contact-rich manipulation
~\cite{calandra2018more,xue2025reactivediffusionpolicy}. These methods demonstrate that
current tactile observations improve contact perception and action
control. Vision-based tactile sensors further show that local contact
geometry, shear, and force-related information can be recovered from
deformable sensing surfaces~\cite{yuan2017gelsight,ma2019dense},
motivating tactile representations that preserve more than binary
contact state.

A closely related line of work uses tactile prediction as a learning
signal. TacForeSight predicts short-horizon tactile dynamics from
tactile observations and wrist force/torque signals, and uses the
predicted states as anticipatory contact information
~\cite{zang2026tacforesight}. DreamTacVLA predicts future tactile
signals to improve contact-aware action learning, while
$\mathcal{N}_0$-VTLA introduces a predictive tactile pathway into a
large-scale vision-tactile-language-action model
~\cite{ye2025dreamtacvla,neote2026n0vtla}. In a related direction,
$\tau$ uses future visual observations to supervise the learning of a
spatiotemporal tactile representation~\cite{cheng2026tau}. Other
methods develop tactile or visuo-tactile world models for future-state
prediction, rollout generation, policy learning, and closed-loop
control
~\cite{higuera2026visuotactilewm,huang2026vitacworld,
zheng2026omnivta,zhou2026touchworld,wu2026tactilewam,tian2026vtwam}.
Together, these studies show that both current tactile conditioning
and predictive tactile learning can benefit contact-rich manipulation.

TacWAM builds on this broader visuo-tactile learning direction rather
than treating tactile prediction itself as the primary novelty. It
focuses on how tactile futures are represented and how they are used
inside WAM training: the SAF Encoder and tactile history support
prediction of force and deformation changes, while AGT Attention keeps
future sensory targets from becoming direct inputs to action
generation.

%% file: sec/3_method.tex
\section{Method}

\begin{figure*}[t]
\centering
\includegraphics[width=\textwidth]{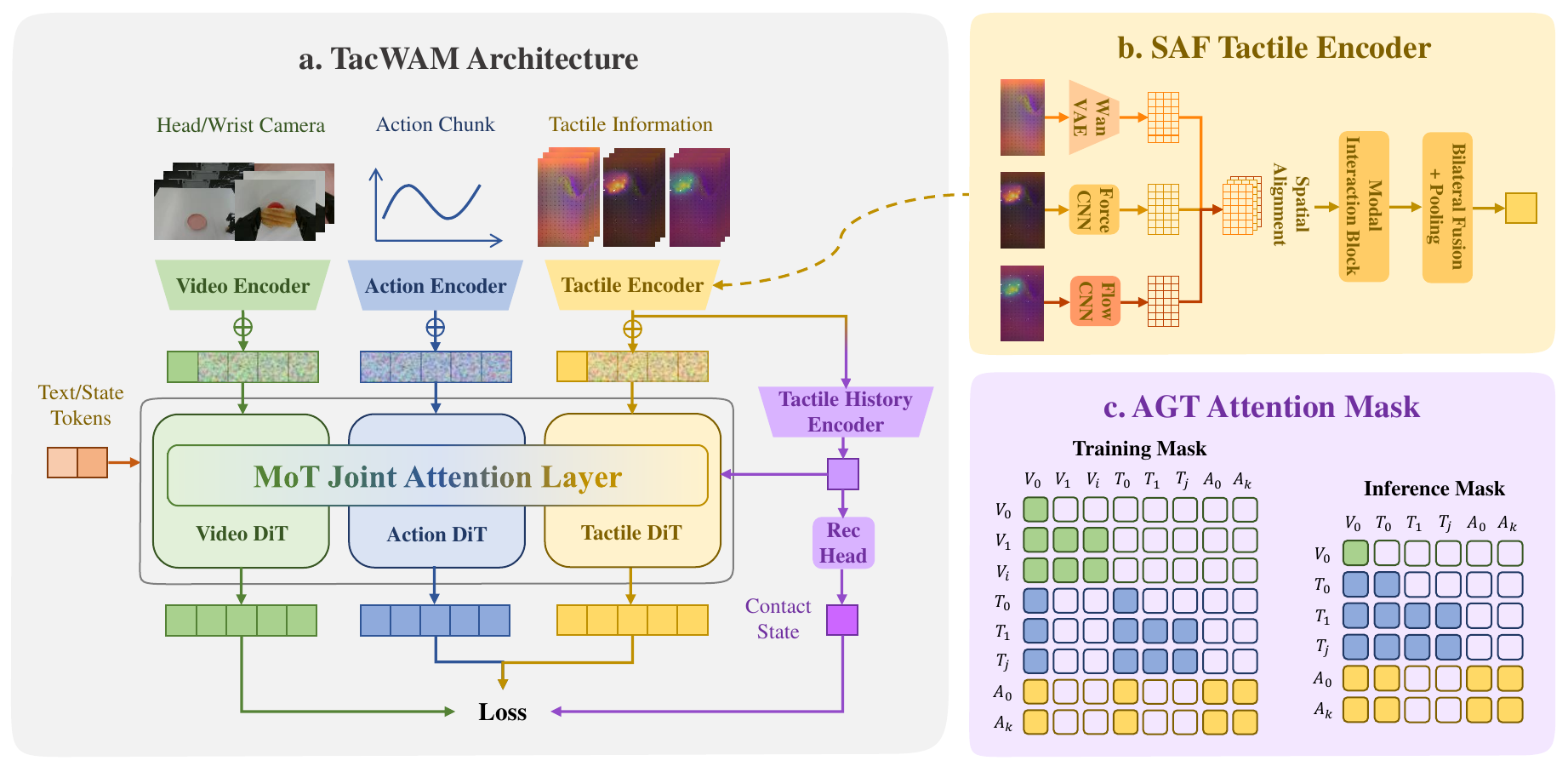}
\caption{Overview of TacWAM. (a) TacWAM jointly predicts visual futures, future tactile states, and action chunks from deployment-available context. (b) The Spatially Aligned Fusion (SAF) Tactile Encoder constructs the tactile latent prediction space from tactile appearance, dense force fields, and mesh deformation flow, with global wrench supervision. (c) Anchor-Guided Tri-Modal (AGT) Attention separates observable anchors, future prediction targets, and action tokens.}
\label{fig:method}
\end{figure*}

\subsection{Problem Formulation}

TacWAM extends a visual World Action Model with predictive tactile state modeling. At time $t$, the robot observes
\begin{equation}
o_t = (o_t^v, o_t^{\mathrm{tac}}, q_t, l),
\end{equation}
where $o_t^v$ is the visual observation, $o_t^{\mathrm{tac}}$ is the tactile observation, $q_t$ is proprioception, and $l$ denotes the task or language condition. Raw tactile readings are first mapped into a learned tactile representation,
\begin{equation}
z_t^{\mathrm{tac}} = E_{\mathrm{SAF}}(o_t^{\mathrm{tac}}).
\end{equation}
Note that we do not assume that $z_t^{\mathrm{tac}}$ is the full physical state. It is a tactile-observable representation of force- and deformation-related signals, used as the prediction space for future tactile modeling.

At the beginning of an action chunk $t_0$, TacWAM predicts visual futures, future tactile states, and an action sequence from stream-specific information. Here $c_{t_0}^{\mathrm{tac}}$ denotes a compact tactile history context summarized from recent tactile latents; its construction is detailed in the Tactile History-Modulated Prediction subsection.
\begin{equation}
\begin{aligned}
\hat{\mathbf v}_{t_0+1:t_0+H}
&= F_V(o_{t_0}^v, q_{t_0}, l),\\
\hat{\mathbf z}^{\mathrm{tac}}_{t_0+1:t_0+H}
&= F_T(o_{t_0}^v, z_{t_0}^{\mathrm{tac}}, c_{t_0}^{\mathrm{tac}}, q_{t_0}, l),\\
\hat{\mathbf a}_{t_0:t_0+H-1}
&= F_A(o_{t_0}^v, z_{t_0}^{\mathrm{tac}}, q_{t_0}, l).
\end{aligned}
\end{equation}
These equations summarize the intended conditioning rather than separate networks; noised prediction tokens and the flow-matching timestep are omitted for clarity. Following the underlying WAM, each action token is represented as the target robot state for the next transition, $a_t := q_{t+1}^{\mathrm{target}}$, rather than as a low-level instantaneous command.

\subsection{TacWAM Overview}

Figure~\ref{fig:method}(a) provides an overview of the overall architecture. TacWAM is built around a single tri-modal WAM generator with visual, tactile, and action streams. On top of this generator, TacWAM introduces three tactile-specific designs. First, the SAF Tactile Encoder maps synchronized tactile appearance, force, and deformation-flow signals into a shared tactile latent space, which defines the future tactile prediction target. Second, the tactile history encoder summarizes recent tactile changes before the current chunk and conditions the tactile prediction branch on contact-phase context. Third, AGT Attention specifies the information topology among visual, tactile, and action streams, preventing future sensory prediction targets from being directly accessed by action tokens.

The three streams are jointly implemented in a Mixture-of-Transformers backbone, where visual, tactile, and action streams are processed by modality-specific experts and communicate through masked mixed self-attention. AGT governs this mixed self-attention among visual, tactile, and action tokens; task language and current proprioception are supplied separately as deployment-available context through expert cross-attention. Under the finalized AGT mask, future tactile prediction is a parallel predictive objective rather than an action-conditioned consequence model: the action stream uses deployment-available visual and tactile anchors, task/proprioceptive context, and action-token interactions, while future tactile states provide predictive supervision for force and deformation changes during training.

\subsection{Spatially Aligned Fusion (SAF) Tactile Encoder}

TacWAM defines a tactile latent prediction space using $E_{\mathrm{SAF}}$ and a tactile reconstructor $R_{\mathrm{tac}}$, as illustrated in Figure~\ref{fig:method}(b). Each tactile observation contains synchronized signals from the two gripper-side tactile sensors:
\begin{equation}
o_t^{\mathrm{tac}} =
(I_t^{\mathrm{rect}}, F_t, M_t^{\mathrm{flow}}),
\end{equation}
where $I_t^{\mathrm{rect}}$ is the rectified tactile image, $F_t$ is the dense local force field, and $M_t^{\mathrm{flow}}$ is the mesh deformation flow. These three inputs are spatially registered on the tactile sensor surface and are fused into a step-level tactile representation:
\begin{equation}
z_t^{\mathrm{tac}}
= E_{\mathrm{SAF}}(I_t^{\mathrm{rect}}, F_t, M_t^{\mathrm{flow}}).
\end{equation}

For each sensor side, tactile appearance, force, and deformation-flow features are encoded with corresponding branches, and the two sides are fused bilaterally. The encoder is single-frame so that both observed tactile states and predicted per-step tactile states share the same latent space. To preserve force- and deformation-related structure, $R_{\mathrm{tac}}$ reconstructs dense force fields, mesh deformation flow, and a bilateral resultant wrench from $z_t^{\mathrm{tac}}$. The resultant wrench contains a 3D force and a 3D torque for each gripper-side tactile sensor:
\begin{equation}
R_{\mathrm{tac}}(z_t^{\mathrm{tac}})
= (\hat{F}_t, \hat{r}_t^{\mathrm{wrench}}, \hat{M}_t^{\mathrm{flow}}).
\end{equation}
The resultant wrench is therefore treated as global mechanics supervision rather than a fourth spatial input. TacWAM predicts future $z^{\mathrm{tac}}$ states instead of directly generating heterogeneous tactile sensor streams. The reconstruction objectives encourage the latent state to preserve local force, deformation, and bilateral wrench information.

\subsection{Anchor-Guided Tri-Modal (AGT) Attention}

TacWAM uses AGT to specify which information each stream may read during mixed self-attention, as shown in Figure~\ref{fig:method}(c). At chunk start $t_0$, the current tactile state is used as a clean anchor,
\begin{equation}
T_0 = z_{t_0}^{\mathrm{tac}},
\end{equation}
and the supervised future tactile sequence is
\begin{equation}
\mathbf T_{1:H}
= [z_{t_0+1}^{\mathrm{tac}}, \ldots,
z_{t_0+H}^{\mathrm{tac}}].
\end{equation}
The WAM generator arranges tokens as
\begin{equation}
[V_0, V_{1:H} \mid T_0, T_{1:H} \mid A],
\end{equation}
where $V_0$ is the current visual anchor, $V_{1:H}$ denotes future visual prediction tokens, $T_0$ is the clean, i.e., unnoised, tactile anchor, $T_{1:H}$ denotes future tactile prediction tokens, and $A$ denotes action tokens. Future visual and tactile tokens are noised prediction variables during flow matching, while their clean counterparts provide supervision. We use the notation ``query reads key/value'' to describe visibility. Table~\ref{tab:agt} summarizes the AGT rules.

\begin{table}[t]
\centering
\caption{AGT visibility. Rows denote query streams and entries list permitted key/value streams.}
\label{tab:agt}
\begin{tabular}{lp{0.62\linewidth}}
\toprule
Query stream & Permitted key/value streams \\
\midrule
Visual & $V_0$ and $V_{1:H}$ under the video first-frame-causal mask \\
Tactile future & $V_0$, $T_0$, and $T_{1:H}$ under a tactile first-frame-causal mask \\
Action & $V_0$, $T_0$, and action tokens \\
\bottomrule
\end{tabular}
\end{table}

The central constraint is that action tokens cannot read future visual or tactile prediction tokens. Moreover, video tokens cannot read $T_{1:H}$, and future tactile tokens do not read action tokens. Future visual and tactile states can therefore serve as prediction targets during training, while action generation remains conditioned only on deployment-available information. AGT is thus an information-isolation mechanism for predictive co-training, not a claim of action-conditioned tactile dynamics.

\subsection{Tactile History-Modulated Prediction}

Single-frame touch is often ambiguous: the same force magnitude can correspond to stable contact, growing pressure, slip recovery, or impending object damage depending on the recent interaction trajectory. TacWAM therefore summarizes recent tactile history before the current chunk:
\begin{equation}
c_{t_0}^{\mathrm{tac}}
= E_{\mathrm{hist}}
(z_{t_0-T_{\mathrm{hist}}+1:t_0}^{\mathrm{tac}}).
\end{equation}
The output $c_{t_0}^{\mathrm{tac}}$ is a compact chunk-level tactile context. It is not inserted as additional memory tokens. Instead, it modulates the tactile expert through adaptive normalization, helping future tactile prediction disambiguate the current interaction phase and tactile evolution trend. The history context is not a direct privileged input to the action stream.

The history window and the tactile generation sequence share the current state: the former uses it as the endpoint of recent tactile evolution, while the latter uses it as the clean initial anchor $T_0$ for future prediction. During flow-matching training, future tactile tokens are noised while $T_0$ remains clean. The tactile expert predicts $\hat{\mathbf z}^{\mathrm{tac}}_{t_0+1:t_0+H}$ from the current visual anchor, the clean tactile anchor, tactile history context, and its own tactile prediction sequence. It does not attend to action tokens under the finalized AGT mask.

This design changes the role of tactile prediction. TacWAM does not attach tactile forecasting as an isolated post-hoc head. Instead, future tactile states are integrated into the multimodal generative objective, encouraging the jointly trained WAM to encode contact-relevant tactile evolution beyond visual supervision alone.

\subsection{Training and Inference}

Training first learns the tactile latent prediction space by optimizing $E_{\mathrm{SAF}}$ and $R_{\mathrm{tac}}$. The tactile reconstruction loss combines force-field, bilateral resultant wrench, and mesh-flow reconstruction. After this tactile representation pretraining, the tactile encoder and decoder are frozen.
\begin{equation}
\mathcal L_{\mathrm{SAF}}
= \lambda_F \mathcal L_F
+ \lambda_R \mathcal L_{\mathrm{wrench}}
+ \lambda_M \mathcal L_{\mathrm{flow}} .
\end{equation}

TacWAM is then trained with the frozen tactile representation model. The frozen tactile encoder provides $z^{\mathrm{tac}}$ for tactile history and future tactile targets. The WAM generator is trained to jointly predict visual futures, action chunks, and tactile futures with scheduler-weighted flow-matching objectives. Tactile latent prediction is supervised on $\mathbf T_{1:H}$, and the frozen but differentiable $R_{\mathrm{tac}}$ decodes predicted tactile states to force fields, bilateral resultant wrench, and mesh flow as a semantic consistency regularizer. The decoder parameters remain fixed, while gradients through the decoder supervise the predicted tactile latents. The tactile history encoder is trained jointly with the WAM generator. Here, $\mathcal L_{\mathrm{contact}}$ is a lightweight contact-event auxiliary loss applied to the tactile history encoder. Its labels are derived from the binary contact sequence in the tactile history window, including contact onset, release, and stable-contact states.
\begin{equation}
\begin{aligned}
\mathcal L_{\mathrm{TacWAM}}
&= \lambda_v \mathcal L_{\mathrm{video}}
+ \lambda_a \mathcal L_{\mathrm{action}}
+ \lambda_z \mathcal L_{\mathrm{tac\text{-}latent}} \\
&\quad
+ \lambda_{\mathrm{sem}} \mathcal L_{\mathrm{tac\text{-}decoded}}
+ \lambda_c \mathcal L_{\mathrm{contact}} .
\end{aligned}
\end{equation}

During deployment, the model runs at chunk level and outputs the action sequence from current visual/tactile anchors and task/proprioceptive context. After executing one action chunk, the robot receives new observations and infers the next chunk, yielding a closed-loop receding-horizon execution process. Future tactile predictions can also be decoded through $R_{\mathrm{tac}}$ for analyzing force and deformation evolution, but they are not used as online correction signals in this paper. The main experiments evaluate whether learning these predictive tactile states improves tactile forecasting and action generation under the AGT information constraints.

%% file: sec/4_experiments.tex
\section{Experiments}

We evaluate TacWAM through the following research questions:
\begin{itemize}[leftmargin=*]
    \item RQ1: How does TacWAM compare with vision-only and visuo-tactile baselines on contact-rich manipulation?
    \item RQ2: Does tactile history improve future force prediction and phase-dependent contact manipulation within TacWAM?
    \item RQ3: How do AGT information constraints affect deployment performance compared with relaxed attention masks?
\end{itemize}

\subsection{Experimental Setup}

We evaluate TacWAM on a real-world manipulation platform with synchronized visual, tactile, proprioceptive, and action streams.

\textbf{Platform.}
The robot platform uses an Agilex Piper manipulator equipped with two Xense G1-WS tactile sensors mounted on the gripper sides. Visual observations are captured from an overhead camera and a wrist-mounted camera, providing complementary scene-level and hand-centric views. The robot state and action are both represented by joint positions, and each action token corresponds to a target joint-position state for the next control step. All sensing and control streams are synchronized at 30 Hz.

\textbf{Data.}
The dataset contains demonstrations collected on the real platform, with synchronized visual observations, joint-position states and actions, and tactile streams from the two gripper-side sensors. For each task, we collect 300 episodes, and each episode contains about 500 frames on average. Each tactile frame contains the three spatially aligned inputs used by SAF: rectified tactile appearance, dense force fields, and mesh deformation flow. All sensor streams are temporally aligned at the frame level.

\begin{figure}[t]
\centering
\includegraphics[width=\linewidth]{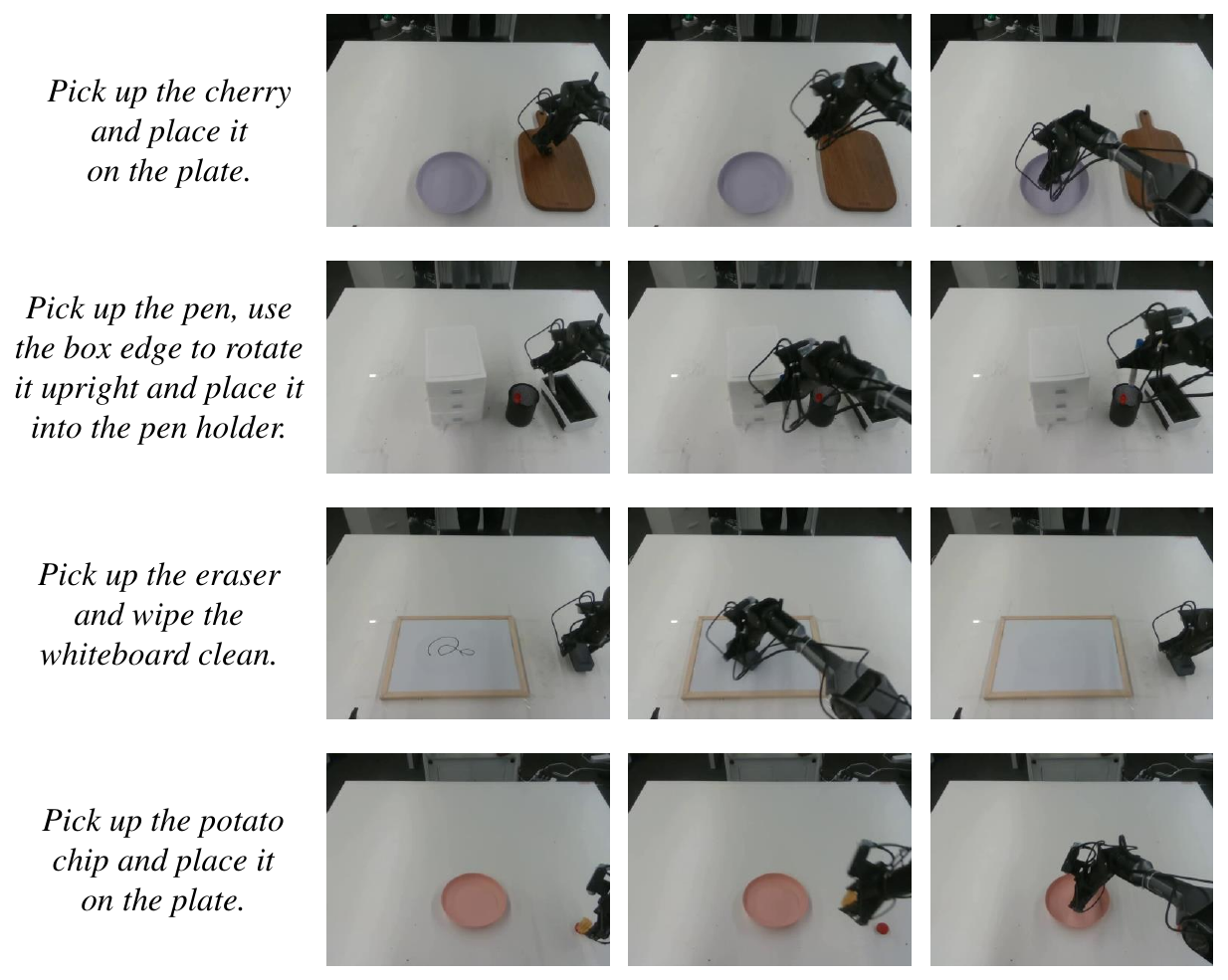}
\caption{Real-world evaluation tasks. The figure shows the text prompts and overhead-camera observations for the four contact-rich manipulation tasks used in our experiments.}
\label{fig:tasks}
\end{figure}

\textbf{Tasks.}
We evaluate on four real-world contact-rich manipulation tasks, which cover fragile grasping, small-object contact localization, sustained contact, and dynamic in-hand manipulation, as shown in Figure~\ref{fig:tasks}:
\begin{itemize}[leftmargin=*]
    \item \textbf{Potato-chip grasping:} grasp a thin, fragile chip and place it onto a target plate, requiring precise force regulation to avoid slip or breakage.
    \item \textbf{Cherry grasping:} grasp an object of variable size, where tactile feedback is needed to confirm stable contact.
    \item \textbf{Whiteboard wiping:} maintain sustained surface contact while moving along the board, stressing pressure control, shear, and contact continuity.
    \item \textbf{Two-pen twirling:} manipulate two pens through dynamic bilateral contact changes, requiring the model to represent evolving force distribution and rotation dynamics.
\end{itemize}
A trial is counted as successful if the task goal is completed without violating the task-specific physical constraint: the chip must be placed without breaking or dropping, the cherry must be stably grasped and placed, the whiteboard must be wiped with sustained contact, and the pens must complete the required twirling motion without falling.

\textbf{Baselines.}
We compare TacWAM with four representative baselines covering vision-only VLA, vision-only WAM, tactile policy learning, and tactile WAM modeling. All methods are trained and evaluated on the same task data and use the same joint-position action representation when applicable. Vision-only baselines use the overhead and wrist camera views together with proprioception and task conditioning, while tactile baselines additionally receive the synchronized tactile observations.
\begin{itemize}[leftmargin=*]
    \item \textbf{$\pi_{0.5}$} is a vision-language-action baseline built on the open-world generalization VLA model from Physical Intelligence~\cite{black2025pi05}. It represents the class of end-to-end visual VLA policies that map visual observations and task instructions directly to robot actions, without explicit visual or tactile future prediction.
    \item \textbf{Fast-WAM} is a vision-only World Action Model baseline~\cite{yuan2026fastwam}. It retains visual predictive co-training during training but removes explicit future generation at inference, providing a visual-predictive WAM comparison for evaluating TacWAM.
    \item \textbf{RDP} is a visual-tactile policy baseline based on Reactive Diffusion Policy~\cite{xue2025reactivediffusionpolicy}. It uses tactile feedback for contact-rich manipulation through a slow-fast policy structure, but it is not a tactile world-action model that predicts future tactile states as WAM training targets.
    \item \textbf{VT-WAM} is a visual-tactile WAM baseline that jointly models visual futures, tactile futures, and actions for contact-rich manipulation~\cite{tian2026vtwam}. Since no official implementation is publicly available, we reimplement VT-WAM following the paper design and train it under the same data and action-space setting.
\end{itemize}

\textbf{Evaluation protocol.}
All methods are trained on the same demonstrations for the same number of epochs under their respective input settings. We evaluate each trained model under the same task definitions, success criteria, and sequence of real-world test conditions, with object poses and initial robot configurations varied within predefined ranges. For each method and task, we run 20 trials and report the success rate as the percentage of successful executions.

\subsection{Main Results}

To answer RQ1, the main evaluation measures whether TacWAM improves closed-loop manipulation success over the four baseline families. Table~\ref{tab:main_success} reports per-task and average success rates. The main comparison focuses on task-level execution performance; diagnostic analyses of tactile prediction quality and architectural components are presented in the next subsection.

\begin{table}[t]
\centering
\caption{Main real-world manipulation results. Entries are task success rates over 20 real-world trials.}
\label{tab:main_success}
\small
\setlength{\tabcolsep}{3pt}
\begin{tabular}{@{}lccccc@{}}
\toprule
Method & Chip & Cherry & Wiping & Twirling & Avg. \\
\midrule
$\pi_{0.5}$ & 10.0\% & 60.0\% & 35.0\% & 20.0\% & 31.3\% \\
Fast-WAM & 15.0\% & 40.0\% & 20.0\% & 5.0\% & 20.0\% \\
RDP & 50.0\% & 45.0\% & 0.0\% & 35.0\% & 32.5\% \\
VT-WAM & 45.0\% & 45.0\% & 30.0\% & 30.0\% & 37.5\% \\
\midrule
TacWAM & \textbf{90.0\%} & \textbf{70.0\%} & \textbf{75.0\%} & \textbf{65.0\%} & \textbf{75.0\%} \\
\bottomrule
\end{tabular}
\end{table}

\textbf{Quantitative analysis.}
TacWAM outperforms all baselines on every task. Compared with the strongest baseline for each task, TacWAM improves success rates by $40.0$ points on potato-chip grasping, $10.0$ points on cherry grasping, $40.0$ points on whiteboard wiping, and $30.0$ points on two-pen twirling. On average, TacWAM achieves a success rate of $75.0\%$, exceeding the best-performing baseline (VT-WAM, $37.5\%$) by $37.5$ percentage points. The gains are especially pronounced on whiteboard wiping and two-pen twirling, the two tasks that most strongly require sustained contact and dynamic bilateral contact reasoning, where purely visual or reactive tactile baselines struggle most.

The overall trend matches our expectation: visual-only baselines are competitive when contact is visually simple, while tactile baselines improve on tasks requiring force regulation, sustained contact, or bilateral contact reasoning. TacWAM further improves over both reactive tactile policies and existing tactile WAMs, supporting the value of mechanics-aware tactile predictive training in the complete framework.

\textbf{Qualitative analysis.}
The incorporation of tactile sensing and predictive tactile training mainly addresses two problems.
First, precise grasping of irregular small objects: on potato-chip and cherry grasping, where the objects are fragile or of variable size, vision-only baselines frequently miss the object entirely or over-squeeze it; in contrast, tactile feedback provides direct evidence of contact and force, leading to substantially higher success than vision-only policies.
Second, under visual occlusion the robot cannot reliably determine whether contact is established or whether the pens are actually rotating. Vision-only models often produce motions that appear visually plausible but are physically ineffective, for example when the wiper glides over the whiteboard without maintaining contact. Tactile observations provide information that is unavailable from vision, allowing TacWAM to maintain sustained contact and handle dynamic bilateral interaction more reliably. The force-rollout analysis below further shows that the predicted resultant force magnitude follows the ground-truth trend over complete contact-rich episodes.

\subsection{Ablation Results}

\begin{figure}[t]
\centering
\includegraphics[width=0.85\linewidth]{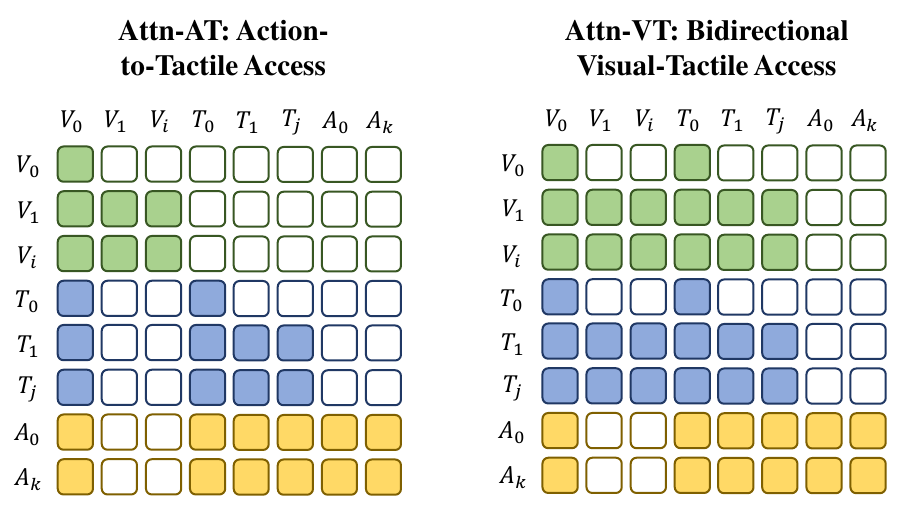}
\caption{Attention-mask variants used in the joint-training ablations. Attn-AT allows action tokens to read the full tactile prediction sequence $T_{0:H}$. Attn-VT includes Attn-AT and further enables bidirectional interaction between visual future tokens and tactile future tokens.}
\label{fig:mask_ablation}
\end{figure}

To answer RQ2 and RQ3, we use three joint-training ablations to analyze the roles of tactile history and AGT visibility. These ablations form a nested sequence that progressively removes history and relaxes cross-stream restrictions, while keeping the same task data, action representation, and tactile future prediction objective. This staged design tests whether performance degrades consistently as the proposed temporal context and information constraints are weakened. Figure~\ref{fig:mask_ablation} illustrates the two relaxed attention masks used after removing tactile history. Table~\ref{tab:ablation_success} reports success rates on two representative contact-rich tasks.

\begin{table}[t]
\centering
\caption{Ablation results on selected contact-rich tasks. Entries are task success rates over 20 real-world trials.}
\label{tab:ablation_success}
\begin{tabular}{lccc}
\toprule
Variant & Chip & Wiping & Avg. \\
\midrule
TacWAM & \textbf{90.0\%} & \textbf{75.0\%} & \textbf{82.5\%} \\
\midrule
w/o History & 50.0\% & 60.0\% & 55.0\% \\
w/o History + Attn-AT & 30.0\% & 45.0\% & 37.5\% \\
w/o History + Attn-VT & 10.0\% & 5.0\% & 7.5\% \\
\bottomrule
\end{tabular}
\end{table}

\textbf{Without tactile history.}
This ablation removes the tactile history pathway to test whether $E_{\mathrm{hist}}$ and the resulting history context are useful for contact-rich manipulation. The history encoder is not used, the tactile history context is replaced by zeros, and the auxiliary contact-event loss $\mathcal L_{\mathrm{contact}}$ is disabled. All other joint-training objectives remain unchanged, including future tactile prediction, decoded tactile targets, visual prediction, and action prediction. Removing tactile history reduces the average success rate from $82.5\%$ to $55.0\%$ ($-27.5$ points), with a particularly large drop on chip grasping ($90.0\% \rightarrow 50.0\%$), where the force buildup preceding stable grasping is difficult to infer reliably from the current tactile anchor alone. This result suggests that temporal tactile context provides information beyond the current tactile anchor and is an important part of TacWAM.

\textbf{Without tactile history + Attn-AT.}
This ablation starts from the history-free variant and relaxes only the action-stream tactile visibility. Instead of allowing action tokens to read only the clean tactile anchor $T_0$, action tokens can read the full tactile prediction sequence $T_{0:H}$ together with $V_0$ and other action tokens. The video mask, tactile first-frame-causal mask, and tactile access to the visual anchor are otherwise unchanged. Performance drops further to $37.5\%$ on average. Rather than helping, exposing action tokens to future tactile prediction tokens creates an unreliable shortcut: during training, action tokens can exploit target-derived future prediction variables, whereas deployment relies on generated counterparts under the inference process. This train--test mismatch degrades deployment behavior, suggesting that the restriction $A \leftarrow T_0$ is important for TacWAM.

\textbf{Without tactile history + Attn-VT.}
This ablation further relaxes the information structure by enabling direct bidirectional interaction between visual prediction tokens and tactile future tokens, while preserving the clean-anchor semantics. Specifically, the current visual anchor can read the tactile anchor, future visual tokens can read tactile prediction tokens, the tactile anchor can read the current visual anchor, and future tactile tokens can read visual prediction tokens. The tactile stream still keeps its first-frame-causal internal structure. This variant nearly collapses, reaching only $7.5\%$ on average ($-30.0$ points from Attn-AT) and $5.0\%$ on whiteboard wiping. Fully bidirectional video--tactile future exchange appears to destabilize multimodal co-training and severely degrades action generation. This result supports AGT's stricter separation for deployment-consistent learning.

Overall, the three ablations form a consistent trend: performance decreases as tactile history is removed and the attention structure is progressively relaxed. These staged results provide consistent evidence that tactile history and restricted future-token visibility are important to the complete TacWAM design.

\begin{figure}[t]
\centering
\includegraphics[width=0.9\linewidth]{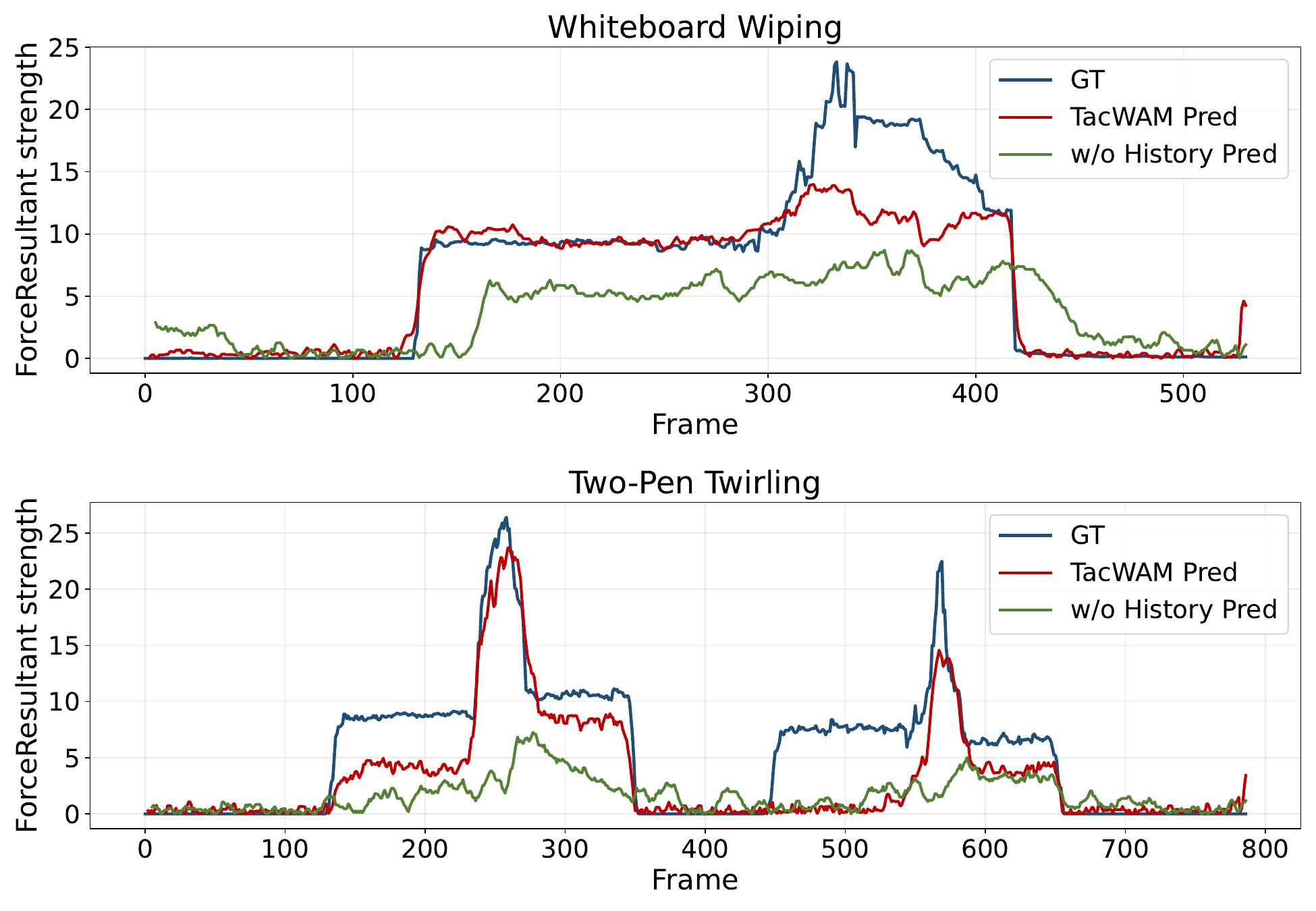}
\caption{Representative episode-level tactile prediction rollouts on whiteboard wiping and two-pen twirling. We reconstruct the bilateral resultant wrench from predicted tactile states, compute the resultant force magnitude, and compare TacWAM and w/o History predictions with ground truth over complete episodes.}
\label{fig:tactile_prediction}
\end{figure}

\textbf{Future tactile state prediction.}
For RQ2, we further evaluate whether tactile history improves predicted force evolution over complete episodes. For whiteboard wiping and two-pen twirling, we decode predicted tactile states to reconstruct the bilateral resultant wrench and compute the resultant force magnitude. Figure~\ref{fig:tactile_prediction} compares the resulting curves from TacWAM and the history-free variant against ground truth. TacWAM tracks both the magnitude and temporal trend of the ground-truth force more closely than w/o History, illustrating that tactile history can improve future tactile prediction rather than only improving downstream action success.

The difference is especially clear on whiteboard wiping, where the w/o History prediction lags behind the ground-truth force changes. This temporal delay suggests that, without recent tactile context, the model needs additional incoming tactile evidence before updating its estimate of the current contact phase. By contrast, tactile history provides recent interaction evolution before the current chunk, allowing TacWAM to infer the current contact state more promptly and predict force changes with less delay. This analysis explains why removing history substantially reduces manipulation success in the ablation results.

%% file: sec/5_conclusion.tex
\section{Conclusion}
We presented TacWAM, a tactile-augmented World Action Model that extends visual WAMs with predictive tactile states for contact-rich manipulation. The SAF Encoder constructs a mechanics-aware tactile representation from spatially aligned tactile appearance, dense force fields, and deformation flow with global wrench supervision; tactile history summarizes recent force and deformation changes; and AGT Attention enables tri-modal co-training with explicit future-target isolation. On four contact-rich manipulation tasks, TacWAM outperforms vision-only VLA/WAM and visuo-tactile baselines on every task, improving the average success rate by $37.5$ percentage points over the strongest baseline. Staged ablations show consistent degradation when tactile history is removed and attention restrictions are progressively relaxed, supporting the importance of tactile history and controlled information flow. These results suggest that predictive tactile states are a useful training signal for contact-rich world action modeling.


%% file: sec/6_appendix.tex
\appendix

This supplementary material provides implementation, training, hardware,
data-collection, evaluation, ablation, and baseline details for TacWAM.
It is intended to support reproducibility and reviewer inspection of design
choices that are summarized in the main paper.
Future tactile states are treated as predictive training targets rather than
direct action inputs; decoded future tactile predictions are used for analysis
only and are not online correction signals at deployment.

\section{Implementation Details}

\subsection{Overview of TacWAM Implementation}

Figure~\ref{fig:method} summarizes the main TacWAM components.
TacWAM is implemented as a tri-modal Mixture-of-Transformers (MoT) World
Action Model with visual, tactile, and action experts.
The experts communicate through masked mixed self-attention governed by
Anchor-Guided Tri-Modal (AGT) Attention.
Task language and current proprioception are provided as
deployment-available context through expert cross-attention.

At the start of an action chunk $t_0$, TacWAM jointly predicts:
\begin{itemize}[leftmargin=*]
    \item visual futures $\hat{\mathbf v}_{t_0+1:t_0+H_v}$,
    \item future tactile latents $\hat{\mathbf z}^{\mathrm{tac}}_{t_0+1:t_0+H}$,
    \item an action chunk $\hat{\mathbf a}_{t_0:t_0+H-1}$.
\end{itemize}
Future visual and tactile tokens are noised prediction variables under
flow matching.
Under the finalized AGT mask, future tactile prediction is a parallel
predictive objective rather than an action-conditioned consequence model:
the action stream uses only deployment-available anchors, task/proprioceptive
context, and action-token interactions.

Table~\ref{tab:arch_dims} lists the main architectural dimensions used in
our implementation.

\begin{table}[t]
\centering
\caption{Architectural dimensions used by TacWAM.}
\label{tab:arch_dims}
\small
\begin{tabular}{lc}
\toprule
Component & Value \\
\midrule
Tactile latent dimension $D_{\mathrm{ctrl}}$ & $128$ \\
History context dimension & $256$ \\
History length $T_{\mathrm{hist}}$ & $32$ \\
Action / tactile horizon $H$ & $32$ \\
Video frames per chunk (after subsampling) & $9$ \\
Action / proprioception dimension & $7$ \\
Video expert hidden dim / layers / heads & $3072$ / $30$ / $24$ \\
Action expert hidden dim / layers / heads & $1024$ / $30$ / $24$ \\
Tactile expert hidden dim / layers / heads & $1024$ / $30$ / $24$ \\
Text context length & $128$ \\
\bottomrule
\end{tabular}
\end{table}

\subsection{SAF Tactile Encoder and Decoder}

The Spatially Aligned Fusion (SAF) tactile encoder $E_{\mathrm{SAF}}$ is a
strictly single-frame encoder.
Each tactile observation contains synchronized signals from the two
gripper-side sensors,
\begin{equation}
o_t^{\mathrm{tac}}
=
\bigl(I_t^{\mathrm{rect}}, F_t, M_t^{\mathrm{flow}}\bigr),
\end{equation}
with the following tensor shapes used in our implementation:
\begin{itemize}[leftmargin=*]
    \item rectified tactile appearance $I_t^{\mathrm{rect}}\in\mathbb{R}^{2\times 3\times H\times W}$,
    \item dense force field $F_t\in\mathbb{R}^{2\times 35\times 20\times 3}$,
    \item mesh deformation flow $M_t^{\mathrm{flow}}\in\mathbb{R}^{2\times 35\times 20\times 3}$.
\end{itemize}
The second dimension indexes the left and right tactile sensors.
In our real-robot pipeline, rectified tactile images are resized to
$H\times W = 224\times 128$.

For each sensor side, three branches encode appearance, force, and
deformation flow:
\begin{itemize}[leftmargin=*]
    \item Appearance uses a frozen Wan video VAE encoder, producing a
    $48$-channel spatial latent grid that defines the common spatial
    reference.
    \item Dense force and mesh flow are encoded by shared convolutional
    grid encoders (output channel widths $32$ each) and bilinearly aligned
    to the appearance latent grid.
    \item Per-side fused features are obtained by concatenating the three
    branch latents ($48{+}32{+}32$) and applying convolutional fusion
    (fused channel width $64$).
\end{itemize}
Left and right fused maps are then concatenated and passed through a
bilateral fusion block, followed by global spatial pooling and an MLP
control readout that yields a $128$-dimensional tactile latent
$z_t^{\mathrm{tac}}$.
Left/right branch encoders share weights so that both sensors map into a
common latent space.

The tactile reconstructor $R_{\mathrm{tac}}$ is a single-step multi-head
decoder with a shared MLP trunk
($128\!\rightarrow\!256\!\rightarrow\!256$) and three reconstruction heads:
\begin{itemize}[leftmargin=*]
    \item dense force field $\hat{F}_t\in\mathbb{R}^{2\times 35\times 20\times 3}$,
    recovered by projecting to a $64\times 7\times 5$ feature map,
    bilinearly upsampling to $35\times 20$, and applying convolutions;
    \item mesh deformation flow $\hat{M}_t^{\mathrm{flow}}\in\mathbb{R}^{2\times 35\times 20\times 3}$,
    using the same spatial-head structure;
    \item bilateral resultant wrench
    $\hat{r}_t^{\mathrm{wrench}}\in\mathbb{R}^{12}$,
    consisting of a $3$D force and a $3$D torque for each of the two
    gripper-side sensors.
\end{itemize}
The bilateral resultant wrench is used only as global mechanics
supervision; it is not a fourth spatial encoder input.
Reconstruction targets are normalized with training-set statistics
(per-channel / per-dimension $z$-score, with clipping for wrench and flow
targets) before SmoothL1 supervision.
TacWAM predicts future $z^{\mathrm{tac}}$ states rather than directly
generating heterogeneous raw tactile streams.

\subsection{Tactile History Encoder}

Single-frame touch can be ambiguous across contact phases.
TacWAM therefore summarizes a fixed recent window of teacher tactile
latents before the current chunk:
\begin{equation}
c_{t_0}^{\mathrm{tac}}
=
E_{\mathrm{hist}}
\bigl(z_{t_0-T_{\mathrm{hist}}+1:t_0}^{\mathrm{tac}}\bigr).
\end{equation}
Implementation defaults are listed in Table~\ref{tab:arch_dims}.
The encoder is a lightweight temporal residual Conv1d network:
\begin{itemize}[leftmargin=*]
    \item input projection $128\!\rightarrow\!256\!\rightarrow\!256$ with SiLU,
    \item $3$ residual temporal blocks (LayerNorm, depthwise Conv1d with
    kernel size $3$, SiLU, pointwise linear, residual add),
    \item last / mean / max / std pooling over the $32$-step hidden
    sequence, concatenated to a $1024$-D vector,
    \item global MLP readout $1024\!\rightarrow\!512\!\rightarrow\!256$.
\end{itemize}
The output is a compact chunk-level vector $c_{t_0}^{\mathrm{tac}}$;
history is not inserted as additional attention tokens.
The history window and the tactile generation sequence share the current
state: history uses it as the endpoint of recent tactile evolution, while
generation uses it as the clean initial anchor $T_0$.

\subsection{Adaptive Normalization for History Modulation}

The history context modulates only the tactile expert.
Concretely, $c_{t_0}^{\mathrm{tac}}$ is projected by an MLP
($256\!\rightarrow\!1024\!\rightarrow\!6\times 1024$) to a modulation vector
and added to the tactile expert's flow-matching timestep modulation before
DiT blocks ($D_{\mathrm{hidden}}=1024$).
This adaptive-normalization pathway conditions future tactile prediction
on recent interaction evolution without exposing history tokens to the
action stream as privileged inputs.

\subsection{Contact-Event Auxiliary Loss}

A lightweight contact-event head is attached to the history encoder and
trained with binary contact labels derived from the history window.
From a $32$-step binary contact sequence, we construct a $6$-dimensional
target covering:
(1)~current contact,
(2)~any contact in the window,
(3)~contact onset,
(4)~contact release,
(5)~stable contact
(current contact with recent contact duration $\ge 8$),
and
(6)~stable no-contact
(current no-contact with recent no-contact duration $\ge 8$).
The auxiliary loss is BCE-with-logits with weight
$\lambda_c=0.1$.
Contact labels are derived offline from thresholded resultant-force
magnitude computed from the tactile streams. The history head predicts only
these binary contact-event labels; raw force and wrench vectors are
supervised through SAF reconstruction and decoded tactile consistency, not
through $\mathcal L_{\mathrm{contact}}$.
In the history-ablated variant, the history encoder is disabled,
$c_{t_0}^{\mathrm{tac}}$ is replaced by zeros, and
$\mathcal L_{\mathrm{contact}}$ is removed.

\subsection{Action/State Representation and Chunked Inference}

Robot state and action are both represented by $7$-DoF joint positions.
Following the underlying WAM formulation, each action token denotes the
target robot state for the next transition,
$a_t := q_{t+1}^{\mathrm{target}}$, rather than an instantaneous low-level
command.
Actions and proprioceptive states are normalized with dataset $z$-score
statistics.

Chunk geometry used in training and deployment:
\begin{itemize}[leftmargin=*]
    \item action / tactile prediction horizon $H=32$,
    \item video sampling uses $33$ temporally aligned frames with an
    action--video frequency ratio of $4$, yielding $9$ video frames per
    chunk (current frame plus $8$ future visual prediction frames),
    \item tactile future tokens are aligned to the action horizon, with the
    first tactile token kept as a clean current anchor during training
    (TI2V-style tactile anchoring: noisy future tokens are overwritten at
    index $0$ by the clean teacher latent).
\end{itemize}
Camera images are captured at $480\times 640$ and resized to $224\times 224$
per view; the two views are concatenated horizontally for the video expert.
At deployment, TacWAM runs in a receding-horizon loop: it executes one
action chunk, receives new observations, and infers the next chunk.
Default inference uses $20$ flow-matching denoising steps.
Future tactile predictions may be decoded through $R_{\mathrm{tac}}$ for
force/deformation analysis, but they are not used as online correction
signals in this work.

\section{Attention Mask Details}

Tokens are arranged as
\begin{equation}
[V_0, V_{1:H_v} \mid T_0, T_{1:H} \mid A],
\end{equation}
where $V_0$ is the current visual anchor, $V_{1:H_v}$ denotes future visual
prediction tokens, $T_0$ is the clean tactile anchor,
$T_{1:H}$ denotes future tactile prediction tokens, and $A$ denotes action
tokens.
We use ``query reads key/value'' to describe visibility.
Boolean attention entries are True when the query may attend to the
corresponding key/value.
Figure~\ref{fig:mask_ablation} illustrates the AGT mask and the two relaxed
ablation masks.

\subsection{Full AGT Mask Definition}

Under the finalized AGT mask:
\begin{itemize}[leftmargin=*]
    \item \textbf{Visual $\rightarrow$ Visual.}
    Video tokens follow a first-frame-causal mask within the visual stream:
    the current visual frame is an anchor, and future visual tokens attend
    under the video expert's first-frame-causal rule.
    Visual tokens do not read tactile tokens.
    \item \textbf{Tactile $\rightarrow$ Visual / Tactile.}
    Future tactile tokens may read the current visual anchor $V_0$
    (first-frame visual tokens only).
    Within the tactile stream, a first-frame-causal mask is applied:
    $T_0$ attends only to itself, while later tactile tokens may attend to
    the full tactile sequence $T_{0:H}$.
    Future tactile tokens do not read action tokens.
    \item \textbf{Action $\rightarrow$ Visual / Tactile / Action.}
    Action tokens may read $V_0$, $T_0$, and all action tokens
    (full action-to-action visibility).
    Action tokens cannot read future visual tokens or future tactile
    prediction tokens $T_{1:H}$.
\end{itemize}
Table~\ref{tab:agt} summarizes the AGT visibility rules.
Table~\ref{tab:agt_matrix} gives a compact stream-level visibility matrix
(``Y'' = permitted under the corresponding within-stream causal rule;
``--'' = blocked).

\begin{table}[t]
\centering
\caption{Stream-level AGT visibility matrix. Entries indicate whether the
query stream may read the key/value stream under the within-stream causal
rules described above.}
\label{tab:agt_matrix}
\small
\begin{tabular}{lccccc}
\toprule
Query $\backslash$ KV & $V_0$ & $V_{1:H_v}$ & $T_0$ & $T_{1:H}$ & $A$ \\
\midrule
Visual & Y & Y & -- & -- & -- \\
Tactile & Y & -- & Y & Y & -- \\
Action & Y & -- & Y & -- & Y \\
\bottomrule
\end{tabular}
\end{table}

\subsection{Attn-AT Mask}

The Attn-AT ablation starts from the history-free variant and relaxes only
action-stream tactile visibility.
Action tokens may read the full tactile prediction sequence $T_{0:H}$
together with $V_0$ and other action tokens.
The video mask, tactile first-frame-causal mask, and tactile access to
$V_0$ remain unchanged.

\subsection{Attn-VT Mask}

The Attn-VT ablation further enables bidirectional interaction between
visual prediction tokens and tactile future tokens while preserving
clean-anchor semantics:
\begin{itemize}[leftmargin=*]
    \item the current visual anchor may read the tactile anchor,
    \item future visual tokens may read tactile prediction tokens,
    \item the tactile anchor may read the current visual anchor,
    \item future tactile tokens may read visual prediction tokens.
\end{itemize}
The tactile stream still keeps its first-frame-causal internal structure,
and action tokens retain Attn-AT visibility ($V_0$, $T_{0:H}$, and action
tokens).

\subsection{Query/Key-Value Visibility Clarification}

AGT is an information-isolation mechanism for predictive co-training, not a
claim of action-conditioned tactile dynamics.
In particular:
\begin{itemize}[leftmargin=*]
    \item future tactile states supervise the tactile prediction branch;
    \item action generation remains conditioned only on deployment-available
    information under AGT;
    \item future tactile tokens do not attend to action tokens, so tactile
    futures are not modeled as action-conditioned consequences in the
    finalized design.
\end{itemize}

\subsection{Train--Test Mismatch for Attn-AT}

Allowing action tokens to read $T_{0:H}$ creates an unreliable training
shortcut.
During training, action tokens can exploit target-derived future prediction
variables (noised teacher tactile futures whose clean counterparts provide
supervision), whereas deployment-time generation must rely on model-generated
counterparts under the inference process.
This target-derived shortcut is absent under deployment-time generation,
creating a train--test mismatch that degrades closed-loop behavior. This is
why AGT restricts action visibility to $A\leftarrow T_0$.

\section{Training Details}

Training proceeds in two stages.
We report final hyperparameters only (no search ranges).

\subsection{Tactile Representation Pretraining}

Stage~1 jointly trains $E_{\mathrm{SAF}}$ and $R_{\mathrm{tac}}$ with
\begin{equation}
\mathcal L_{\mathrm{SAF}}
=
\lambda_F \mathcal L_F
+
\lambda_R \mathcal L_{\mathrm{wrench}}
+
\lambda_M \mathcal L_{\mathrm{flow}},
\end{equation}
where each term is SmoothL1 on normalized targets.
Table~\ref{tab:stage1_loss} lists the final Stage~1 loss weights.
The Wan VAE appearance encoder remains frozen.
Final Stage~1 optimization settings are given in
Table~\ref{tab:train_hp}.
After Stage~1, the tactile encoder and decoder are frozen for TacWAM
training.

\begin{table}[t]
\centering
\caption{Stage~1 SAF reconstruction loss weights.}
\label{tab:stage1_loss}
\small
\begin{tabular}{lc}
\toprule
Term & Weight \\
\midrule
Dense force field $\lambda_F$ & $1.0$ \\
Bilateral resultant wrench $\lambda_R$ & $1.5$ \\
Mesh deformation flow $\lambda_M$ & $1.0$ \\
\bottomrule
\end{tabular}
\end{table}

\subsection{TacWAM Training with Frozen Tactile Representation}

Stage~2 freezes $E_{\mathrm{SAF}}$ and $R_{\mathrm{tac}}$, then trains the
tri-modal WAM generator and the tactile history encoder.
Frozen $E_{\mathrm{SAF}}$ provides $z^{\mathrm{tac}}$ for history windows
and future tactile targets.
Frozen but differentiable $R_{\mathrm{tac}}$ decodes predicted tactile
latents to force fields, bilateral resultant wrench, and mesh flow as a
semantic consistency regularizer; decoder parameters remain fixed while
gradients through the decoder supervise predicted latents.

The joint objective is
\begin{equation}
\begin{aligned}
\mathcal L_{\mathrm{TacWAM}}
&=
\lambda_v \mathcal L_{\mathrm{video}}
+
\lambda_a \mathcal L_{\mathrm{action}}
+
\lambda_z \mathcal L_{\mathrm{tac\text{-}latent}} \\
&\quad
+
\lambda_{\mathrm{sem}} \mathcal L_{\mathrm{tac\text{-}decoded}}
+
\lambda_c \mathcal L_{\mathrm{contact}},
\end{aligned}
\end{equation}
with final weights in Table~\ref{tab:stage2_loss}.
Video, action, and tactile-latent terms follow scheduler-weighted MSE under
flow matching; decoded tactile consistency uses MSE on normalized Stage~1
targets and excludes the clean tactile anchor frame.
Tactile-latent and decoded-tactile losses are likewise computed only on
future indices $1{:}H$, not on the clean anchor.

\begin{table}[t]
\centering
\caption{Stage~2 TacWAM loss weights.}
\label{tab:stage2_loss}
\small
\begin{tabular}{lc}
\toprule
Term & Weight \\
\midrule
Video flow matching $\lambda_v$ & $1.0$ \\
Action flow matching $\lambda_a$ & $1.0$ \\
Tactile latent flow matching $\lambda_z$ & $1.0$ \\
Decoded tactile consistency $\lambda_{\mathrm{sem}}$ & $0.5$ \\
Contact-event auxiliary $\lambda_c$ & $0.1$ \\
\bottomrule
\end{tabular}
\end{table}

\begin{table}[t]
\centering
\caption{Final training hyperparameters for Stage~1 and Stage~2.}
\label{tab:train_hp}
\small
\begin{tabular}{lcc}
\toprule
Setting & Stage~1 & Stage~2 \\
\midrule
Optimizer & AdamW & AdamW \\
Learning rate & $1\times 10^{-4}$ & $1\times 10^{-4}$ \\
LR schedule & cosine & cosine \\
Weight decay & $1\times 10^{-2}$ & $1\times 10^{-2}$ \\
Batch size (per GPU) & $32$ & $8$ \\
Gradient accumulation & $1$ & $2$ \\
Epochs & $20$ & $20$ \\
Mixed precision & bfloat16 & bfloat16 \\
Max gradient norm & $1.0$ & $1.0$ \\
Flow-matching timesteps & -- & $1000$ / stream \\
Scheduler shift (train/infer) & -- & $5.0$ \\
Inference denoising steps & -- & $20$ \\
Fixed random seed & $42$ & $42$ \\
\bottomrule
\end{tabular}
\end{table}

Backbone initialization follows the Wan2.2-TI2V-5B MoT setup.
Action and tactile experts are initialized from an ActionDiT checkpoint
aligned to the Wan2.2 backbone (hidden size $1024$).
Each method uses one final checkpoint.
We do not report training-seed variance; real-robot variation comes from
$20$ evaluation trials per task.

\subsection{Ablation Training Protocol}

The staged ablations in the main paper are trained under the same Stage~2
data, action representation, horizons, and future-tactile prediction
objective, with the following controlled changes:
\begin{itemize}[leftmargin=*]
    \item \textbf{w/o History:}
    disable the tactile history encoder; replace
    $c_{t_0}^{\mathrm{tac}}$ by zeros; disable
    $\mathcal L_{\mathrm{contact}}$; keep AGT.
    \item \textbf{w/o History + Attn-AT:}
    start from the history-free setting and set action visibility to
    $A\leftarrow\{V_0,T_{0:H},A\}$.
    \item \textbf{w/o History + Attn-VT:}
    start from Attn-AT and enable bidirectional video--tactile future
    visibility while preserving first-frame-causal anchor semantics.
\end{itemize}
All ablation variants use the same optimization hyperparameters as full
TacWAM Stage~2 (Table~\ref{tab:train_hp}).

\section{Hardware and Software Environment}

\subsection{Robot Platform and Sensors}

\begin{itemize}[leftmargin=*]
    \item Robot platform: Agilex Piper manipulator.
    \item Tactile sensors: two Xense G1-WS sensors mounted on the gripper
    sides.
    \item Cameras: one overhead camera and one wrist-mounted camera
    (raw resolution $480\times 640$, resized to $224\times 224$).
    \item Synchronization: visual, tactile, proprioceptive, and action
    streams are temporally aligned at $30$\,Hz.
    \item Action/state interface: $7$-DoF joint-position targets.
\end{itemize}

\subsection{Compute and Software}

Training and inference use the following software stack
(from the project dependency specification):
\begin{itemize}[leftmargin=*]
    \item Python $\ge 3.10$,
    \item PyTorch $2.7.1$ (+CUDA $12.8$ wheels),
    \item torchvision $0.22.1$ (+CUDA $12.8$ wheels),
    \item accelerate $1.12.0$,
    \item DeepSpeed $0.18.5$ (ZeRO stage used for multi-GPU TacWAM training),
    \item transformers $4.49.0$,
    \item hydra-core $1.3.2$,
    \item numpy $1.26.4$.
\end{itemize}
Hardware details for the machines used in the reported experiments:
\begin{itemize}[leftmargin=*]
    \item GPU model: NVIDIA A800-SXM4-80GB ($80$\,GB),
    \item number of GPUs used for TacWAM Stage~2 training: $8$,
    \item CPU model: Intel Xeon Platinum 8374C @ $2.70$\,GHz
    ($2$ sockets, $124$ logical CPUs),
    \item system RAM: $1.8$\,TB,
    \item operating system: Ubuntu $22.04.5$ LTS,
    \item NVIDIA driver: $535.129.03$,
    \item CUDA toolkit: $12.8$ (matching the PyTorch CUDA $12.8$ build).
\end{itemize}

\section{Data Collection Details}

We collect real-robot demonstrations on the platform above for four
contact-rich tasks:
potato-chip grasping, cherry grasping, whiteboard wiping, and two-pen
twirling.
For each task:
\begin{itemize}[leftmargin=*]
    \item number of demonstrations: $300$ episodes,
    \item average episode length: about $500$ frames,
    \item streams: synchronized visual observations (overhead and wrist),
    dual gripper-side tactile readings (rectified appearance, dense force
    fields, and mesh deformation flow), proprioception, and joint-position
    actions,
    \item sampling rate: $30$\,Hz frame-level alignment,
    \item action/state representation: joint positions.
\end{itemize}
Raw data are not publicly released due to custom hardware calibration,
platform-specific synchronization, and real-robot data management
constraints.
Publicly available alternatives do not provide the same synchronized dual
tactile sensing, visual observations, joint-state/action representation,
and contact-rich task setup used in this study.

\subsection{Code and Artifact Release}

Upon publication, TacWAM training and inference code, including model
definitions, losses, and attention-mask configurations, will be released
under a research-use license.
Raw data, preprocessing scripts, analysis scripts, trained checkpoints, and
baseline reimplementation code are not released.

\section{Evaluation Protocol}

All methods are trained on the same demonstrations for the same number of
epochs under their respective input settings and evaluated under the same
protocol:
\begin{itemize}[leftmargin=*]
    \item $20$ real-robot trials per method per task,
    \item identical task definitions, success criteria, and test-condition
    sequence across methods,
    \item object poses and robot initial configurations varied within
    predefined ranges,
    \item receding-horizon execution: execute one predicted action chunk,
    receive a new observation, then infer the next action chunk,
    \item reported metric: success rate $=$
    (number of successful executions) / $20$.
\end{itemize}
We do not report confidence intervals or statistical significance tests in
this work.
Each compared method uses one final checkpoint; observed variation reflects
real-robot trial outcomes rather than multi-seed retraining.

\section{Task-Specific Success Criteria}

A trial is counted as successful only if the task goal is completed without
violating the corresponding physical constraint:
\begin{itemize}[leftmargin=*]
    \item \textbf{Potato-chip grasping:}
    the chip is grasped and placed onto the target plate without breaking
    or dropping.
    \item \textbf{Cherry grasping:}
    the cherry is stably grasped and placed without dropping or visible
    crushing.
    \item \textbf{Whiteboard wiping:}
    the eraser maintains sustained contact with the board and the target
    region is wiped to a predefined cleanliness standard.
    \item \textbf{Two-pen twirling:}
    the required twirling motion is completed without the pens falling.
\end{itemize}

\section{Baseline Details}

All baselines are trained and evaluated on the same task data and use the
same joint-position action representation when applicable.
Vision-only baselines receive overhead and wrist views with proprioception
and task conditioning; tactile baselines additionally receive synchronized
tactile observations.
Each baseline uses one final checkpoint and is evaluated with $20$ trials
per task under the protocol in Section~6.

\begin{itemize}[leftmargin=*]
    \item \textbf{$\pi_{0.5}$}~\cite{black2025pi05}:
    a vision-language-action baseline that maps visual observations and
    task instructions directly to robot actions, without explicit visual or
    tactile future prediction.
    We fine-tune the $\pi_{0.5}$ model family on each task using AdamW,
    cosine decay with peak learning rate $1\times 10^{-4}$,
    batch size $64$, action horizon $33$, EMA decay $0.999$,
    mixed precision bfloat16, and $20$ training epochs.
    \item \textbf{Fast-WAM}~\cite{yuan2026fastwam}:
    a vision-only World Action Model baseline that retains visual predictive
    co-training during training but removes explicit future generation at
    inference.
    We train Fast-WAM with the same MoT backbone family and chunk geometry as
    TacWAM, without tactile streams, using AdamW, learning rate
    $1\times 10^{-4}$, cosine schedule, weight decay $1\times 10^{-2}$,
    batch size $4$, gradient accumulation $2$, $20$ epochs, and bfloat16.
    One final checkpoint is used for evaluation.
    \item \textbf{RDP}~\cite{xue2025reactivediffusionpolicy}:
    a visual-tactile Reactive Diffusion Policy baseline with a slow--fast
    structure that uses tactile feedback for contact-rich control, but does
    not predict future tactile states as WAM training targets.
    We reimplement RDP following the paper design under our shared data and
    joint-position action setting, using the same synchronized visual,
    tactile, proprioceptive, and task-conditioning inputs available to
    tactile policy baselines.
    \item \textbf{VT-WAM}~\cite{tian2026vtwam}:
    a visual-tactile WAM baseline that jointly models visual futures,
    tactile futures, and actions.
    No official public implementation is available; we reimplement VT-WAM
    following the paper design and train it under the same data and
    action-space setting. The reimplementation uses the same synchronized
    visual, tactile, proprioceptive, and task-conditioning inputs as TacWAM,
    while following VT-WAM's visual-tactile-action prediction design rather
    than TacWAM's history-modulated AGT formulation.
\end{itemize}

\section{Additional Qualitative Results}

Figure~\ref{fig:tactile_prediction} provides additional episode-level force-rollout
visualizations consistent with the main-paper analysis.
Predicted tactile states are decoded to bilateral resultant wrench and
summarized by resultant force magnitude.
These curves are diagnostic analyses of predictive tactile training; they
are not used as online action-correction signals.

No additional invented quantitative results are reported beyond the
main-paper success rates and the qualitative prediction comparisons above.